\documentclass[]{article}
\usepackage{geometry}
\usepackage{amsfonts}
\usepackage{amsmath}
\usepackage{tikz}
\usepackage{cite}
\usepackage{algorithm}
\usepackage{algpseudocode}
\usetikzlibrary{shapes.geometric, arrows, positioning}

\title{Tutorial on the Probabilistic Unification of Estimation Theory, Machine Learning, and Generative AI}
\author{
	Mohammed S. Elmusrati\thanks{School of Technology and Innovations, University of  Vaasa, Finland email: mohammed.elmusrati@uwasa.fi}
}

\date{}

\begin{document}

\maketitle

\begin{abstract}

Extracting meaning from uncertain, noisy data is a fundamental problem across time series analysis, pattern recognition, and language modeling. This survey presents a unified mathematical framework that connects classical estimation theory, statistical inference, and modern machine learning, including deep learning and large language models. By analyzing how techniques such as maximum likelihood estimation, Bayesian inference, and attention mechanisms address uncertainty, the paper illustrates that many AI methods are rooted in shared probabilistic principles. Through illustrative scenarios—including system identification, image classification, and language generation—we show how increasingly complex models build upon these foundations to tackle practical challenges like overfitting, data sparsity, and interpretability. In other words, the work demonstrates that maximum likelihood, MAP estimation, Bayesian classification, and deep learning all represent different facets of a shared goal: inferring hidden causes from noisy and/or biased observations. It serves as both a theoretical synthesis and a practical guide for students and researchers navigating the evolving landscape of machine learning.

\end{abstract}

\section{Introduction}
Let us begin with a classical and straightforward estimation problem, described by the following model:

\begin{equation}
	y = x + n,
\end{equation}

where \( y \) denotes the observed measurement, \( x \) is the unknown parameter to be estimated, and \( n \) represents zero-mean random measurement noise.

In the absence of any prior information about \( x \), the most suitable approach is the \textbf{maximum likelihood estimator (MLE)}, which is derived  as \cite{elmu}:

\begin{equation}
	\hat{x} = \frac{1}{N} \sum_{i=1}^{N} y_i,
\end{equation}

where \( y_i \) is the \( i \)-th measurement, and \( N \) is the total number of observations.

However, when prior information about \( x \) is available, the MLE is no longer optimal. For example, if it is known a priori that \( x > 0 \), this information can be incorporated into the estimation process to improve accuracy. One way to do this is through the \textbf{maximum a posteriori (MAP)} estimation, which combines the observed data with the prior distribution of \( x \) to produce a more informed and accurate estimate.

This basic estimation problem can be generalized to a more complex and realistic setting as:

\begin{equation}
	\label{gene}
	y_t = f(\mathbf{x}, t) + n_t,
\end{equation}

where \( \mathbf{x} \) is the input vector, \( y_t \) is the observation and \( n_t \) is an unknown en measurement noise or bias. The function or mapping \( f(\cdot, t) \) may describe a static or time-dependent system. It can be also vector of functions and then we will have a vector of outputs or observations \( \mathbf{y}_t\). This formulation is particularly relevant in domains such as finance and control, where the data and underlying models are often represented as time series. Temporal patterns and dependencies in such data can be exploited through time correlations.

The general form presented in Equation~(\ref{gene}) encompasses a wide range of applications, including dynamic modeling, system identification, machine learning, and even generative AI  such as the case of  large language models. By leveraging the structure and dynamics encoded in the model, it is possible to achieve improved estimation and prediction across diverse domains. In what follows, we explore a few  different   scenarios for addressing the uncertainties represented by Equation~(\ref{gene}) and discuss their potential applications.

\begin{itemize}
	\item \textbf{Scenario 1: Inferring Hidden Causes from Observations.}  
	In this scenario, we are given the observation vector $\mathbf{y}_t$, and the objective is to estimate the hidden systematic cause $\mathbf{x}$. This setting is prevalent in numerous fields. For instance, $\mathbf{y}_t$ could represent observed symptoms in healthcare, vibration patterns in electric motors, or fluctuations in the stock prices of a particular company, among countless other real-world applications. The underlying assumption is that these observations are driven by latent factors $\mathbf{x}$, which are mapped to $\mathbf{y}_t$ through a (possibly unknown) static or dynamic function $f(\cdot,t)$.  	
	Importantly, the observations are typically corrupted by an unknown noise component, which may represent measurement noise or biases—either intentional or unintentional. The observation vector $\mathbf{y}_t$ can take various forms. It may be continuous-valued, such as measurements of vibration amplitude or product price. Alternatively, it may be discrete, encoding class membership. For example, $\mathbf{y}_t$ can be a $K$-dimensional vector where the $k$th component lies in the interval $[0,1]$, indicating the likelihood that the input $\mathbf{x}$ belongs to class $k$.\\
	In some cases of system modeling or identification, the function $f(\cdot,t)$ can be estimated based on conceptual understanding or physical modeling. In such scenarios, the functional form of the model is known, while the model parameters remain unknown. There are two possible sub-cases to consider:
	
	\textbf{1. Known Input Scenario:}  
	It is possible to apply a designed input $\mathbf{x}$ and measure the resulting output $\mathbf{y}_t$. This setup facilitates the estimation of the model parameters, often through optimization techniques. A typical application of this approach is in wireless channel modeling, where the transmitter sends a predefined signal (often called a pilot or training sequence), and the receiver uses the corresponding output to estimate the channel parameters. By accurately estimating these parameters, we can mitigate the negative impact of the wireless channel on subsequent data transmissions using inverse modeling techniques, commonly referred to as equalizers. However, since wireless channels are typically dynamic and time-varying, these pilot symbols must be transmitted repeatedly—often every few milliseconds—which reduces spectral efficiency.
	
	\textbf{2. Unknown Input Scenario:}  
	In this more challenging case, we neither control nor know the system input $\mathbf{x}$, yet we still aim to optimize the model parameters. This situation arises in many real-world systems where direct excitation is impractical or impossible. Despite the absence of known inputs, parameter estimation techniques—often based on statistical inference, expectation-maximization, or unsupervised learning—can still be employed to approximate the underlying model.
	\item \textbf{Scenario 2: Inferring the Mapping Using Causes and Observations.} 
	
	This second scenario corresponds to what is commonly known today as \textit{machine learning} or \textit{data-driven modeling}. In this case, the function—or more generally, the mapping—$f\left(\cdot,t\right)$ can vary from a simple linear relationship to an extremely complex, potentially intractable function. It may be as straightforward as linear regression, or as sophisticated as a deep neural network with numerous hidden layers and millions of interconnected neurons.
	
	The mapping can be either \textit{static}, where the relationship is purely input-output based—such as in image recognition tasks like distinguishing between dogs and cats—or \textit{dynamic}, where the output depends not only on the current input but also on previous inputs. Dynamic mappings arise in applications involving sequential data, such as time series signals or written text, where models like recurrent neural networks (RNNs) are typically employed.
	
	In this scenario, we utilize the available input-output dataset to learn both the structure and the parameters of the mapping function $f(\cdot,t)$. Thus, the learning process consists of determining an appropriate model structure and tuning its parameters to fit the data. In simpler cases, the structure of the function can be predefined, and only its parameters are optimized using the training data.
	
	The simplest case is the static linear model, expressed as:
	\begin{equation}
		\label{LinEq}
		y_k = a_0 + a_1 x_{1,k} + a_2 x_{2,k} + \dots + a_M x_{M,k}
	\end{equation}
	The data in this context consist of $M$  input features and a single output variable. The learning process involves optimizing the parameters $\left(a_0, \dots, a_M\right)$ to minimize a cost function that quantifies the discrepancy between the predicted and actual outputs. Once training is complete, the resulting model is validated using a separate dataset to assess its generalization performance.
	This simple linear model is particularly desirable when it performs well, as it is both easy to train and highly interpretable. For example, when the input features are normalized, the learned parameters directly reflect the relative importance of each feature. Consider a task where the goal is to estimate the probability that a person will develop diabetes within five years based on their medical history. If $x_i$ represents the Body Mass Index (BMI), then the corresponding coefficient $a_i$ quantifies the relative influence of BMI on the risk of developing diabetes.
	
	Unfortunately, simple linear regression models often fail to perform well on complex recognition tasks, as the underlying relationship between inputs and outputs is typically far more intricate than a linear approximation can capture. This limitation is known as the \textit{underfitting problem}, where the model is too simplistic to represent the true structure of the data.
	
	Hence, more sophisticated methods—such as multi-layer neural networks and deep learning algorithms—can be employed to achieve significantly higher mapping capabilities between the input and output. These advanced techniques are capable of modeling highly nonlinear and complex relationships within the data.
	
	However, most of these methods are often referred to as \textit{black-box} models because their internal parameters and decision-making processes are difficult to interpret in relation to the original input features. This lack of interpretability poses challenges, especially in critical applications where understanding the rationale behind a prediction is essential.
	
	Moreover, using a model that is excessively complex relative to the true underlying relationship between input and output can lead to the problem of \textit{overfitting}. In such cases, the model may simply memorize the training data rather than learn to generalize from it. As a result, the model performs well on the training set but fails to capture the true pattern and performs poorly on unseen data. 
	
	This issue can typically be identified by evaluating the learned model on a separate validation dataset, where a significant drop in performance compared to the training set indicates overfitting.
	
\item \textbf{Scenario 3: Large Language Models} 
Language is fundamentally a spoken form of communication, while written language serves as a symbolic representation of spoken phonemes. As such, speech is inherently a time series signal, exhibiting a highly complex internal structure and underlying logic.

One of the most advanced approaches to developing large language models (LLMs)—capable of generating text, answering questions, and performing tasks that emulate human cognitive abilities—relies on training these models using vast and diverse corpora of digital text, including books, academic papers, articles, and novels. These models undergo autoregressive generation processes that enable them to capture intricate patterns, semantic relationships, and contextual dependencies within large-scale textual data.

Fundamentally, a modern LLM operates according to a predictive function similar to the one described in Equation\~(\ref{gene}). In this context, let $\mathbf{x}$ represent a sequence of $m$ words, and the model's objective is to predict the $(m+1)^{th}$ word. It is important to note that the next word prediction does not depend solely on the immediate preceding word (i.e., the $m^{th}$ word), but rather on a broader context involving multiple prior words. This allows the model to preserve the logical flow and coherence of the language.

Moreover, not all preceding words contribute equally to the prediction. Certain words carry significantly more weight in determining the meaning and direction of the sentence. The importance of each word is influenced by its impact on the overall context—specifically, how much it alters the interpretation or continuity of the sentence. Identifying and leveraging these key contextual signals is a fundamental capability of attention-based architectures used in state-of-the-art LLMs \cite{ref3}.
\end{itemize}
Building on the previous discussion, we demonstrated that the general dynamic form of mapping presented in Equation~(\ref{gene}) can serve as a foundational framework for a wide range of applications, spanning from system identification to large language models. 
While this model includes many additional details, we omit them here to maintain the paper's length at a reasonable level. Readers interested in system modeling are referred to \cite{ref1}, \cite{ref2}, and \cite{ref10} .

\section{Optimal Solution: Probabilistic Approach}

The most abstract way to analyze the problem presented in Equation~(\ref{gene}) is through the lens of uncertainty. For instance, given a measurement or observation $\mathbf{y}$, what can be inferred about the hidden cause $\mathbf{x}$? For simplicity, we have omitted the temporal index $t$; however, the results can later be generalized to incorporate temporal dependencies. Alternatively, the problem can be framed as identifying the optimal mapping function $f(\cdot)$ that yields accurate predictions of the output $\mathbf{y}$ for a given input $\mathbf{x}$. This framework can also be reformulated in an autoregressive manner, such that the input at the next time step is the output of the current step, i.e., $\mathbf{x}_{t+1} = \mathbf{y}_t$. \\
We know that uncertainty about any quantity can be modeled using a probability density function (PDF) in the continuous case, or a probability mass function (PMF) in the discrete case. In the following analysis, we will focus on the continuous case using the PDF framework. However, it is worth noting that for the discrete case, one can incorporate Dirac delta functions into the integral formulation to effectively transform the analysis from integration to summation, thereby accommodating the PMF representation.\\
If we have a random vector $\mathbf{x}$ whose uncertainty is described by a multivariate probability density function (PDF) $p_X(\mathbf{x})$, we can attempt to summarize this uncertainty using a representative or "best" vector. There are several common statistical measures used for this purpose:

\begin{itemize}
	\item Mean (Expected Value): The mean vector $\mathbf{x}_{mean}$ represents the center of mass of the probability distribution. It minimizes the expected squared distance to all realizations of the random vector. Mathematically, it is defined as:
	\begin{equation}
		\textbf{x}_{mean} = \textbf{E}[\mathbf{x}] = \int_{\mathbb{R}^n} \textbf{x} , p_X(\textbf{x}) , d\textbf{x}
	\end{equation}
	\item Median: The median vector $\mathbf{x}_{med}$ is defined such that half of the probability mass lies on each side of it. In the univariate case, it satisfies:
	\begin{equation}
		\int_{-\infty}^{x_{med}} p_X(x) \, dx = \int_{x_{med}}^{\infty} p_X(x) \, dx = \frac{1}{2}
	\end{equation}
	For multivariate distributions, the definition of the median is more complex and typically involves minimizing the sum of distances (e.g., in the sense of the $L_1$ norm) between the median and all realizations of the random vector.
	
	\item Mode: The mode vector $\mathbf{x}_{mode}$ is the most probable value, i.e., the point at which the PDF achieves its maximum:
	\begin{equation}
		\mathbf{x}_{mode} = \arg\max_{\mathbf{x}} \ p_X(\mathbf{x})
	\end{equation}
	It corresponds to the value of $\mathbf{x}$ that is most likely to occur, minimizing the distance to the most probable realization.
\end{itemize}

Each of these measures captures a different aspect of centrality in the probability distribution, and their values may differ depending on the shape and symmetry of the distribution. In symmetric probability distributions, such as the Gaussian (Normal) distribution, statistical measures converge to a single value, resulting  in identical estimates for mean, median, and mode. This characteristic is 
a direct consequence of the distribution's inherent symmetry, which 
eliminates the need for separate parameter estimation.\\
Returning to Equation~(\ref{gene}), any changes in the independent input vector \(\mathbf{x}\) are expected to induce corresponding changes in the dependent output vector \(\mathbf{y}\). Statistically, this implies the existence of a cross-correlation between \(\mathbf{x}\) and \(\mathbf{y}\), the nature of which is governed by the mapping function \(f(\cdot)\). The noise term \(\mathbf{n}\) is typically assumed to be an uncorrelated zero mean Gaussian random variable.

The joint probability distribution of the input and output vectors is denoted by \(p_{X,Y}(\mathbf{x}, \mathbf{y})\), while their statistical relationship can be characterized through the conditional distribution \(p_{Y \mid X}(\mathbf{y} \mid \mathbf{x})\), which represents the probability distribution of the output \(\mathbf{y}\) given a specific input \(\mathbf{x}\).

It is important to note that this statistical formulation remains valid regardless of the specific form of the mapping between the input and output. The only essential requirement is the existence of statistical dependence between \(\mathbf{x}\) and \(\mathbf{y}\). In the absence of such dependence—i.e., if \(\mathbf{x}\) and \(\mathbf{y}\) are uncorrelated or statistically independent—then the conditional distribution reduces to the marginal distribution: \(p_{Y \mid X}(\mathbf{y} \mid \mathbf{x}) = p_Y(\mathbf{y})\). This means that changes in \(\mathbf{x}\) do not affect \(\mathbf{y}\), and hence do not alter our belief or certainty about the output.

As an illustrative example, suppose that the probability \(\mathbb{P}(\mathbf{y} \leq \mathbf{y}_0) = p_0\) remains constant regardless of the value of \(\mathbf{x}\), or that the mean of \(\mathbf{y}\) is invariant with respect to \(\mathbf{x}\). In such cases, we conclude that \(\mathbf{x}\) and \(\mathbf{y}\) are statistically independent or at least uncorrelated. Consequently, no information about \(\mathbf{x}\) can be inferred by observing \(\mathbf{y}\).

However, in many real-world problems, our primary interest lies in the hidden variable \(\mathbf{x}\), and we seek to infer it based on the observed data \(\mathbf{y}\). This motivates the need for the posterior distribution \(p_{X \mid Y}(\mathbf{x} \mid \mathbf{y})\), which describes how our belief about \(\mathbf{x}\) is updated given the observation \(\mathbf{y}\). Unfortunately, computing this posterior is often analytically intractable, especially when the mapping function is unknown and \(\mathbf{x}\) is latent.

Fortunately, a foundational result in statistical inference and uncertainty quantification—Bayes' theorem—provides a principled way to compute this posterior distribution:

\begin{equation}
	\label{Bays}
	p_{X \mid Y}(\mathbf{x} \mid \mathbf{y}) = \frac{p_{Y \mid X}(\mathbf{y} \mid \mathbf{x}) \, p_X(\mathbf{x})}{p_Y(\mathbf{y})}
\end{equation}

This powerful theorem forms the basis of Bayesian analysis and allows us to update our prior belief \(p_X(\mathbf{x})\) about the hidden variable in light of new observational evidence \(\mathbf{y}\).
\section{Estimation of the Hidden Variable}

The optimal estimator of the hidden variable \(\mathbf{x}\) depends on the selected criterion, as discussed in the previous section. In this context, estimation should be performed using the conditional probability distribution given in Equation~(\ref{Bays}).

If the objective is to minimize the mean squared error (MSE), the optimal estimator is the posterior mean, denoted as \(\mathbf{x}_{MMSE}\):
\begin{equation}
	\mathbf{x}_{MMSE} = \mathbb{E}_{X \mid Y}[\mathbf{x} \mid \mathbf{y}] = \int \mathbf{x} \, p_{X \mid Y}(\mathbf{x} \mid \mathbf{y}) \, d\mathbf{x}
\end{equation}

To minimize the expected absolute error (i.e., the \(L_1\)-norm of the estimation error), the posterior median of \(p_{X \mid Y}(\mathbf{x} \mid \mathbf{y})\) serves as the optimal estimator.

On the other hand, minimizing the worst-case error (or maximum estimation error) leads to the maximum a posteriori (MAP) estimator, given by:
\begin{equation}
	\mathbf{x}_{MAP}= \arg\max_{\mathbf{x}} \, p_{X \mid Y}(\mathbf{x} \mid \mathbf{y})
\end{equation}

In the special case where the posterior distribution \(p_{X \mid Y}(\mathbf{x} \mid \mathbf{y})\) is symmetric and unimodal, these three estimators (mean, median, and MAP) coincide. Otherwise, each estimator is optimal with respect to its own specific criterion. \\
Determining the conditional probability distribution \( p_{X \mid Y}(\mathbf{x} \mid \mathbf{y}) \) is a highly challenging task, particularly when the relationship between the observed variables and the underlying parameters is complex or unknown. In practice, several sub-optimal approaches approximate this relationship by introducing a simplified \textit{mapping function}. While such assumptions can significantly reduce the complexity of solving for the hidden parameter, they also introduce bias or oversimplification, potentially limiting the model’s ability to capture the true dependency structure between the hidden variables and the observations.

For instance, if we assume that the uncertainty in the hidden parameter follows a Gaussian distribution, the observation noise is also Gaussian, and the mapping function is linear, then the resulting conditional distribution remains Gaussian. Under these idealized conditions, the optimal estimator is the mean of the distribution, which also coincides with the median and the mode. However, such assumptions rarely hold in real-world scenarios, where the underlying relationships are typically nonlinear and the data distributions often deviate significantly from Gaussianity.

Despite this, it is still possible to approximate the actual joint distribution closely using more expressive models, such as deep learning architectures. Regardless of the specific learning algorithm employed, the learning process can always be framed within the broader context of \textit{statistical inference} \cite{ref2}.

To better illustrate these concepts, consider two learning problems: \textit{image pattern recognition} and \textit{large language models (LLMs)}. Before delving into these examples, let us begin with a simple but insightful example that introduces several key ideas in uncertainty modeling and inference.

\subsection*{A Motivating Example}

Assume a binary classification task where we aim to classify an observation into one of two classes, \( C_1 \) or \( C_2 \), based on two observed features \( x_1 \) and \( x_2 \). The functional relationship between the attributes and the class label is unknown. However, for the purpose of this example, suppose the true decision rule is given by the following expression:

\begin{equation}
	\label{eq:example}
	y = \mathrm{sign}\left( \alpha x_1 + \beta x_2 \sin(\pi x_1) - 1.5 x_1 x_2 + n \right)
\end{equation}

where \( \alpha \) and \( \beta \) are parameters, and \( n \) represents additive noise reflecting the modeling error.

If the exact form of the relationship (as given in Equation~(\ref{eq:example})) were known in advance, the learning task would reduce to a parameter estimation (or optimization) problem: using available data to identify the optimal values of \( \alpha \) and \( \beta \). However, in realistic scenarios, we typically \textit{do not know} the true form of the mapping. Instead, we attempt to learn it implicitly, relying on uncertainty modeling rather than an explicit mathematical formulation.

In our example, we simulate data using \( \alpha = 1 \) and \( \beta = 1 \). The learning strategy is to approximate the class-conditional distribution of the data by assuming a \textit{2D multivariate Gaussian distribution}. This involves estimating the distribution’s parameters—mean vector and covariance matrix—based on the observed samples. We then apply Bayes’ theorem to classify new, previously unseen inputs:

\begin{equation}
	p(y \mid \mathbf{x}) = \frac{p(\mathbf{x} \mid y) \, p(y)}{p(\mathbf{x})}
	\label{eq:bayes}
\end{equation}

The pseudo-code for this process is provided in Algorithm (\ref*{alg}). Using this method, the average classification accuracy over thousands of test samples was approximately \textbf{80\%}. This moderate performance reflects the \textit{mismatch between the assumed Gaussian model and the actual data distribution}, which is shaped by a highly nonlinear decision boundary.

Interestingly, when we reduced the nonlinear contribution of the second term by setting \( \beta = 0.05 \) (instead of 1), the classification accuracy increased to approximately \textbf{96\%}, despite the presence of another nonlinear term \( x_1 x_2 \). This demonstrates that even within a simplified modeling framework (e.g., Gaussian assumption), performance can significantly improve when the nonlinearity in the underlying data-generating process is reduced. 

The main takeaway from this example is that \textit{Bayesian statistical inference} enables us to model and reason about complex relationships in data \textit{without requiring explicit knowledge of the underlying functional form}. This is particularly valuable in domains like pattern recognition, where the actual mapping from inputs to outputs is often unknown and cannot be expressed in closed-form.

Nevertheless, a crucial challenge remains: determining a \textit{probability distribution model} that accurately captures the full variability of real-world data. The closer our model aligns with the true distribution, the better the inference and classification performance we can expect.

\begin{algorithm}
	\caption{Bayesian Classification Using Gaussian Distributions}
	\label{alg}
	\begin{algorithmic}[1]
		\Require Number of samples $n$, Number of test samples $k$
		\Ensure Estimated classification accuracy
		
		\State Generate $n$ samples $x \in \mathbb{R}^{2 \times n}$ from standard normal distribution
		\State Compute overall mean vector $M = \text{mean}(x)$ and covariance matrix $Z_t = \text{cov}(x)$
		\State Generate additive Gaussian noise $no \sim \mathcal{N}(0, 0.1^2)$
		\State Compute labels $y = \text{sign}(x_1 + x_2 \cdot \sin(\pi x_1) - 1.5 x_1 x_2 + no)$
		
		\State Separate positive ($y = 1$) and negative ($y = -1$) samples:
		\State \hskip1em $X_p = \{x_i \mid y_i = 1\}$, $X_m = \{x_i \mid y_i = -1\}$
		
		\State Compute class statistics:
		\State \hskip1em Means: $M_p = \text{mean}(X_p)$, $M_m = \text{mean}(X_m)$
		\State \hskip1em Covariances: $Z_p = \text{cov}(X_p)$, $Z_m = \text{cov}(X_m)$
		\State \hskip1em Priors: $P_p = \frac{|X_p|}{n}$, $P_m = 1 - P_p$
		
		\State Initialize success counter $s \gets 0$
		\For{$i = 1$ to $k$}
		\State Generate random test point $X_n \sim \mathcal{N}(0, I)$
		\State Compute class-conditional probabilities:
		\State \hskip1em $p_1 = \mathcal{N}(X_n; M_p, Z_p) \cdot P_p$
		\State \hskip1em $p_2 = \mathcal{N}(X_n; M_m, Z_m) \cdot P_m$
		\State Compute posterior: $pp = \frac{p_1}{p_1 + p_2}$
		\State Predict label: $\hat{y} \gets \begin{cases}
			1 & \text{if } pp \geq 0.5 \\
			-1 & \text{otherwise}
		\end{cases}$
		\State True label: $y_{\text{true}} = \text{sign}(X_{n,1} + 0.05 X_{n,2} \cdot \sin(\pi X_{n,1}) - 1.5 X_{n,1} X_{n,2})$
		\If{$\hat{y} = y_{\text{true}}$}
		\State $s \gets s + 1$
		\EndIf
		\EndFor
		
		\State Compute accuracy: $\text{accuracy} = s / k$
		\State \Return $\text{accuracy}$
	\end{algorithmic}
\end{algorithm}

\subsection{Image Pattern Recognition}
Pattern recognition is typically considered a specific application of the classification process. The main objective is to determine the class to which a given input belongs, out of several possible predefined classes. 
For example, based on certain patient attributes and symptoms, we may classify the disease (i.e., identify which disease class the patient falls into). In power grids, based on recent measurements taken just before a sudden fault, the goal is to identify the type and location of the fault—fault types can be modeled as distinct classes. Similarly, given specific keywords, one may classify an email as spam or not spam. Based on an image, one may determine whether it depicts a cat or a dog. These are just a few examples of the wide range of classification applications, which may involve patterns in images, text, time series signals, and more \cite{ref9}.
In this discussion, we consider the problem of classifying images as either cats or dogs. Let the $i^{{th}}$ image be represented by a 2D image matrix $\mathbf{x}_i$. Let $y$ be a scalar output indicating the classification result. We define $y = +1$ for images that are certainly dogs, and $y = -1$ for images that are certainly cats. That is, $y \in [-1, 1]$ can be interpreted as a certainty score, where values close to $+1$ indicate strong confidence that the image is of a dog, and values close to $-1$ indicate strong confidence that it is of a cat.
For example, if $y = 0.8$ for one image and $y = 0.5$ for another, the classifier is more confident that the first image is a dog compared to the second one. 
For binary classification, a decision threshold can be established such that inputs with scores above the threshold are classified as dogs, and those below are classified as cats.  
If the prior probabilities of the two classes are equal and the dataset is balanced (i.e., equal number of dog and cat images), the optimal threshold is typically set at zero. In this problem, we aim first to find the distribution of the output variable \(y\) given the input image \(\mathbf{x}\), i.e., the conditional probability \(p_{Y \mid X}(y \mid \mathbf{x})\). According to Bayes’ theorem in (\ref{Bays}):
\begin{equation}
	\label{ImagePat}
	p_{Y \mid X}(y \mid \mathbf{x}) = \frac{p_{X \mid Y}(\mathbf{x} \mid y) \, p_Y(y)}{p_X(\mathbf{x})}
\end{equation}
where \(y \in [-1, +1]\). Determining the actual distributions of \(\mathbf{x}\), \(y\), and the conditional distribution \(p_{X \mid Y}(\mathbf{x} \mid y)\) is a highly challenging task.

We begin with modeling the distribution \(p_X(\mathbf{x})\). This distribution represents all possible variations in 2D images that include cats and dogs of various shapes, sizes, poses, and backgrounds. In practice, the space of possible images is extremely large—potentially in the millions—making this a high-dimensional and complex distribution. It captures all the inherent uncertainty and variability in the image data.

Mathematically, one could vectorize each image and apply estimation techniques, such as Maximum Likelihood Estimation (MLE), to determine the parameters of the distribution. However, this requires a prior assumption about the shape of the distribution. A common and practical choice is the multivariate normal (Gaussian) distribution. In this case, we estimate the mean vector and the covariance matrix from the data. If the image has a resolution of \(N \times M\) pixels, then the covariance matrix will be of size \(MN \times MN\), which is computationally infeasible even for moderately sized images. Furthermore, examining a small area consisting of tens or even more pixels often does not provide significant information on its own. Instead, valuable insights emerge when considering the entire image and the relationships between sets of pixels \cite{ref1}.

To overcome this, dimensionality reduction techniques such as Principal Component Analysis (PCA) can be applied. PCA significantly reduces the dimensionality—often retaining as little as 2\% of the original size—while preserving most of the essential information. More advanced approaches, such as Variational Autoencoders (VAEs), have also been widely used in recent years for efficient and effective representation learning \cite{ref4}.

However, in some cases, the assumption of a multivariate normal distribution is not appropriate. For such scenarios, non-parametric methods like histogram estimation or kernel density estimation may yield more accurate representations of \(p_X(\mathbf{x})\) \cite{ref1}.  

It is important to note that the distribution \(p_X(\mathbf{x})\) captures the variability present in both cat and dog images, regardless of their specific features. As such, it represents the uncertainty or diversity within this broader class of images. For comparison, if we construct a similar distribution over human face images (including both male and female faces), we would observe a significantly different mean vector and covariance matrix. This highlights that the parameters of a distribution encode essential characteristics and structural patterns unique to each image class. 
Having such distributions that model the variability of specific image categories—whether human faces, apples, or any other object—is fundamental for generative AI (GenAI). Once this distribution is learned, we can randomly sample patterns from it. These samples can then be fine-tuned to remain within the statistical variance of the original distribution, resulting in the generation of new, realistic patterns. This principle underpins the generation of synthetic images, videos, or text using generative models \cite{ref3},\cite{ref4}.

The same principles extend beyond images and apply to other high-dimensional data domains. With this understanding, the estimation of the conditional probability \(p_{X \mid Y}(\mathbf{x} \mid y)\) becomes more tractable and conceptually clearer. Here we build the multivariate probability function for dogs ($y=+1$) and the cats  ($y=-1$). In other words,  \(p_{X \mid Y}(\mathbf{x} \mid y=+1)\) represents the distribution of dogs pictures.  Let's we have a new picture $\textbf{x}_i$ and we want to automatically decide if it is for cat or dog. From statistical inference, it is logical to measure the similarity of the new picture with both clusters of dogs or cats, i.e., with the likelihood function, as  
\begin{equation}
	\label{ImagePat1}
	p_{Y \mid X}(y=+1 \mid \mathbf{x}_0) = \frac{p_{X \mid Y}(\mathbf{x}_0 \mid y=+1) \, p_Y(y)}{p_X(\mathbf{x}_0)}
\end{equation}
The terms of Equation (\ref*{ImagePat1}) are computed from collected  datasets (i.e., training images). These parameters depend on the proposed probability distribution functions. For instance, when employing a parametric distribution such as the multivariate normal distribution, the parameters consist of the mean vector and the covariance matrix computed from the training images used to construct the model.

The classification of a new image involves first vectorizing it and then applying the same dimensionality reduction technique used during training. After this preprocessing, we compute the likelihood that the image represents a dog based on the learned distribution.

In the case of the multivariate normal distribution, it is common to take the logarithm of both sides of Equation (\ref*{ImagePat1}). This simplification eliminates the exponential component of the Gaussian function and transforms products and quotients into sums and differences, making the expression easier to manipulate. Since the logarithm is a monotonic function—i.e., for any positive numbers $a$ and $b$, if $a \geq b$ then $\log(a) \geq \log(b)$—the location of the maximum remains unchanged. Consequently, for $f\left(x\right) > 0, \ \forall \ x$ in the domain, if
\[
\hat{x} = \arg \max f(x),
\]
then it is always true to say that
\[
\hat{x} = \arg \max \log(f(x)),
\]

It is important to note that in Equation (\ref*{ImagePat1}), the term $p_X(\mathbf{x}_0)$ serves only to normalize the posterior probabilities, ensuring that
\[
p_{Y \mid X}(y = +1 \mid \mathbf{x}_0) + p_{Y \mid X}(y = -1 \mid \mathbf{x}_0) = 1.
\]
However, this normalization does not influence the classification decision, as it is constant across all classes. Therefore, it can be omitted when the goal is to identify the class with the highest posterior probability.

Finally, it is worth highlighting that this probabilistic formulation does not require an explicit mapping function from the image to the class label (e.g., dog or cat). Instead, the model relies solely on the uncertainty representation provided by the multivariate distribution.
 Selecting the optimal probability distribution capable of capturing all relevant details and distinctions between dogs and cats is a challenging task. These differences span a wide range of factors, including size, shape, background, color, orientation, and more. A common starting point is to assume a normal (Gaussian) distribution; however, there is no guarantee that this assumption will yield satisfactory performance across all scenarios.
 
 An alternative is to adopt a non-parametric approach to construct the underlying uncertainty model. While this method can offer greater flexibility, it often requires a large amount of data and computational resources to approximate the complex distribution accurately.
 
 Another approach involves modeling a highly complex mapping function from the input image to a single output, typically +1 for dogs and -1 for cats. In this supervised learning setup, a large set of labeled images is used during the training phase to iteratively adjust the parameters of the mapping function so as to minimize the error between the predicted and actual labels. The resulting model is then evaluated on a separate validation dataset to assess its generalization capability.
 
 In essence, this is the principle behind deep learning: learning a highly flexible mapping function that classifies labeled data by implicitly defining a similarity measure between the input image and the learned internal representations. The model is tuned such that it produces an output close to +1 when the input image resembles those labeled as dogs, and close to -1 when it resembles cats \cite{ref7}.
 
\subsection{Large Language Models}

Human language has historically been the most effective means of communication. It served as the foundation for social and emotional cohesion and the transmission of knowledge, experiences, scientific understanding, and historical events across generations. At its core, language is a phonetic system, consisting of sound waves generated through the coordinated movement of the tongue, lips, and vocal cords to form distinct phonemes. These sounds carry linguistic meaning, which is recognized and interpreted by listeners understanding the language. 
In essence, language is a temporal signal (time series signal), a sequence of auditory events occurring over time. However, due to the ephemeral nature of sound, humans developed writing systems—encoding language into letters, words, texts, and books—to preserve texts over much longer timescales. Written language, while visually represented, still fundamentally reflects its origin as a time-based signal. 
Processing written text does not negate its temporal structure. A text is a sequential arrangement of words that follows strict syntactic and semantic rules to convey meaning—whether explicit or implicit. This has made natural language processing (NLP) and the automatic extraction of meaning from text a central research topic for nearly a century \cite{ref8}.

However, the structural complexity of language has long hindered the development of efficient algorithms capable of processing it. Significant breakthroughs did not occur until the last decade, which saw unprecedented advances in the field.

To understand the source of this complexity, consider the following simplified example: imagine a single page containing only 100 words, where each word consists of 2 to 6 characters, and the language includes only 25 letters. How many unique pages can be constructed under these constraints? The answer is approximately $10^{840}$—a number far beyond human comprehension. 
For comparison, to fill the observable universe (with a diameter exceeding 93 billion light-years) with grains of sand, we would need fewer than $10^{92}$ grains. That number pales in comparison to the combinatorial possibilities of generating a single 100-word page. Of course, most of these word combinations are nonsensical. Linguistic rules and semantic logic drastically reduce the space of valid, meaningful compositions. 
Nevertheless, language remains one of the richest and most complex domains, resisting computational modeling for decades. Each word is composed of characters that exhibit statistical dependencies—knowing one character constrains the probability of the next. The same applies to words: the occurrence of one word influences the likelihood of the following one. Yet, the situation is even more complex. 
The probability of a word does not depend solely on the immediately preceding word but on a sequence of previous words. Importantly, not all preceding words contribute equally: some have a stronger influence on the meaning or context. Furthermore, the location of a word within the context plays a critical role in interpretation, at least for most languages. 
This is why, for many years, designing an algorithm that could understand human language with fluency and accuracy was viewed as science fiction. The turning point came with the advent of deep learning, particularly when mechanisms such as attention were introduced \cite{ref3}. These models were trained on billions of texts and utilized billions of parameters, allowing them to automatically learn the intricate and hidden patterns that govern language. 
The results were astonishing—even to those who developed them. By leveraging deep learning algorithms at massive scale, it became possible to uncover the statistical structure of language with unprecedented accuracy, using principles grounded in uncertainty modeling and probabilistic inference.
In this section, we will present the mathematical foundations underpinning these algorithms, framed through the lens of probability theory and statistical inference. \\

In this mathematical analysis, we treat the text as a sequence of words that follows an unknown probabilistic pattern. Let us revisit Equation~(\ref{gene}). The input $\mathbf{x}$ represents a sequence of words, and the output $\mathbf{y}$ corresponds to the predicted next word(s). 

In the training of large language models (LLMs), the input is typically represented using tokens, which are subword units (including characters, symbols, and whitespace) \cite{ref8}. However, for simplicity, we assume here that $\mathbf{x}$ is a sequence of complete words, and our task is to predict the next word in the sequence. Without loss of generality, we further assume that only one word is predicted at a time. 

We adopt an autoregressive modeling approach, where the predicted word is appended to the existing sequence, and the model is recursively used to predict the next word. This recursive formulation can be interpreted probabilistically. Using Bayes’ theorem, the uncertainty associated with predicting the next word can be expressed as:

\begin{equation}
	\label{LLM1}
	p_{Y \mid X}(y = \text{"Word"}_k \mid \mathbf{x}_0) = \frac{p_{X \mid Y}(\mathbf{x}_0 \mid y = \text{"Word"}_k) \cdot p_Y(y = \text{"Word"}_k)}{p_X(\mathbf{x}_0)}
\end{equation}

This formulation highlights that the problem reduces to a classification task—albeit with a very large number of possible classes (i.e., vocabulary words). The number of candidate words depends on both the language and the specific context of the text, and can easily range into the tens or hundreds of thousands.

However, the challenge is not solely due to the large number of possible outputs, but also due to the sequential nature of language. For instance, consider the input sequence $\mathbf{x} = \{\text{This},\ \text{computer},\ \text{is}\}$. From a purely set-theoretic perspective, one might consider that:
\[
p(\{\text{This},\ \text{computer},\ \text{is}\}) = p(\{\text{is},\ \text{This},\ \text{computer}\}) = p(\{\text{computer},\ \text{This},\ \text{is}\}),
\]
which assumes no importance of word order. However, in natural language, order matters critically. One way to preserve this order is to represent each word as a tuple that includes its position in the sequence. For example:
\[
\mathbf{x} = \{(\text{This},\ 1),\ (\text{computer},\ 2),\ (\text{is},\ 3)\}.
\]

Beyond position, not all words contribute equally to the meaning of the sentence. For contextual weighting, we may associate an attention weight $\alpha_i$ with each word to reflect its relative importance in the sequence. Thus, we extend our representation as:
\[
\mathbf{x} = \{(\text{This},\ 1,\ \alpha_1),\ (\text{computer},\ 2,\ \alpha_2),\ (\text{is},\ 3,\ \alpha_3)\},
\]
where $0 \leq \alpha_i \leq 1$ and $\sum_i \alpha_i = 1$. In this example, the word “computer” likely carries the most semantic weight, and so $\alpha_2$ should be the largest among the three. This kind of representation is foundational in attention mechanisms used in Transformer architectures.
Computing the posterior probability of the next word using Bayes' theorem requires a brute-force search over the training text, which becomes computationally intensive, especially for large corpora. As a result, direct statistical inference is often impractical. Modern deep learning methods—particularly those based on the Transformer architecture—can be seen as implicitly learning and approximating these probabilistic relationships in a more scalable and efficient manner.
Alternatively, Markov models can provide a suboptimal but computationally efficient approximation of the posterior probability. However, their predictive accuracy is typically inferior compared to the current state-of-the-art performance achieved by Transformer-based models.

The Bayesian approach is powerful, but its accuracy is highly dependent on the size and representativeness of the training corpus. When the corpus is too small or unbalanced, the resulting frequency counts become unreliable, undermining the model’s predictive capability.

A significant challenge in this context is data sparsity, particularly the issue of unseen n-grams. For instance, consider a scenario where we aim to predict the fifth word in a sentence based on the preceding four words—i.e., a 5-gram model. If this specific sequence of four words (in the same order) never appears in the training data, its frequency count will be zero. Consequently, the computed likelihood—and thus the posterior probability—will also be zero. This leads to erroneous results, as it fails to account for potentially plausible sequences that were simply absent from the training set. To mitigate this, smoothing techniques are essential. These can be thought of as introducing a controlled amount of “noise” or adjustment to the model to improve its generalization as will be introduced later. 

In modern Natural Language Processing (NLP), although the foundational principles of Bayesian inference remain central—particularly in how prior knowledge is integrated with observed data—these principles are often embedded within more advanced architectures, such as neural networks (e.g., LSTMs and Transformers). These models are capable of capturing long-range dependencies and subtle linguistic patterns far beyond the reach of traditional n-gram models. Nonetheless, the core Bayesian concept of updating beliefs based on evidence continues to underpin the logic of probabilistic language modeling. \\
As an example, let's consider the abstract of this paper. The abstract contains  159  words and there are 117 distinct words. Let's assume the initial event $X=\{\left(modern,1\right), \left(machine,2\right),\left(learning,3\right)\}$.  From the abstract, we want to find the next word in the 4th location that has the maximum probability, i.e.,
\begin{equation}
	y = \arg \max p_{Y\mid X} \left(Y=y\mid X=\{\left(modern,1\right), \left(machine,2\right),\left(learning,3\right)\}\right)
\end{equation}
We check this conditional probability from the learning text. 
\begin{equation}
p_{Y\mid X} \left(Y=y\mid X\right)=\frac{p_{X\mid Y} \left(X\mid Y=y \right) p\left(y\right)}{p\left(X\right)}
\end{equation}

We begin by preprocessing the text: splitting it into tokens using punctuation marks (e.g., commas, periods, dashes) as delimiters, which are subsequently ignored, and converting all words to lowercase to ensure consistency.

The likelihood equation is formulated by computing the probability of observing the first three words, denoted as \textit{X}, given that the fourth word is \textit{y}, i.e., 
\[
p_{X \mid Y}(X \mid Y = y).
\]
The vocabulary extracted from the text consists of 117 distinct words. Our task is to compute the conditional probability for each possible word \( y \).

For instance, if \( y = \{\text{is}\} \), and this sequence does not occur in the training text, then the corresponding conditional probability is zero:
\[
p_{X \mid Y}(X \mid Y = \{\text{is}\}) = 0.
\]
On the other hand, suppose \( y = \{\text{including}\} \) does occur in the context of the preceding three words; we then have:
\[
p_{X \mid Y}(X \mid Y = \{\text{including}\}) = 1.
\]
Hence, the maximum likelihood for 
\[
p_{Y \mid X}(Y = y \mid X)
\]
is achieved when \( y = \text{"including"} \).

Using an autoregressive model, we can generate the fifth and subsequent words based on the learned probability distribution. In this small training corpus, the result tends to reproduce the original sequence of the abstract, such as:
\textit{modern machine learning, including deep learning and large language...}
However, if the given initial sequence of words does not exist in the training text in the same order, the conditional probability becomes zero, making it impossible to predict the next word using maximum likelihood estimation. To address this issue, smoothing techniques are employed to avoid assigning zero probability to unseen events \cite{ref11}. Common methods include \textbf{Laplace smoothing}, which adds a small constant to all counts, and the more advanced \textbf{Kneser-Ney smoothing}, which is particularly effective for language modeling tasks.

Consider the bigram statement \( X = \{ \left(\text{inference}, 1\right), \left(\text{principle}, 2\right) \} \), which does not appear in the training text. Consequently, the likelihood probability becomes zero. Kneser-Ney smoothing is commonly used to handle such cases in $n$-gram models \cite{ref11}. It deals with unseen $n$-grams by backing off to lower-order $n$-grams and adjusting the probabilities accordingly. For example, in a trigram model predicting the third word given the first two, the smoothed probability is defined as:

\begin{equation}
	\label{KN}
	\hat{p}_{Y\mid X}(Y=y \mid X) = \frac{\max\{C(X, y) - d, 0\}}{\sum_y C(X, y)} + \lambda(X) p_{\text{c}}(Y=y \mid X - \{\text{principle}\})
\end{equation}

Where:
\begin{itemize}
	\item \( C(X, y) \) is the count of the trigram "inference principle \( y \)",
	\item \( d \) is the discount parameter,
	\item \( \lambda(X) \) is the backoff weight ensuring that probabilities sum to one,
	\item \( p_{\text{c}}(Y = y \mid X - \{\text{inference}\}) \) is the continuation probability of \( y \) given "principle", after dropping "inference".
\end{itemize}

Since the bigram \textbf{"inference principle"} does not appear in the training text, the first term on the right-hand side of Equation~(\ref{KN}) evaluates to zero. Thus, the Kneser-Ney backoff model simplifies to:

\begin{equation}
	\hat{p}_{Y \mid X}(Y = y \mid X) \approx \lambda(X) \cdot p_{\text{c}}(Y = y \mid X - \{\text{principle}\})
\end{equation}

Now, we aim to predict the most probable word following the word \textbf{"inference"} based on the training text. In our preprocessing, all characters have been converted to lowercase, and punctuation has been removed. The word \textbf{inference} appears twice in the text (the abstract). The total number of distinct bigrams is approximated as \( N = 117 \).

We apply the Kneser-Ney formula as follows:
\begin{equation}
	\hat{p}_{Y \mid X}(Y = y \mid X = \{\text{inference}\})  =
	\frac{\max(C(\{\left(\text{inference},1\right), \left(y,2\right)\}) - d, 0)}{C(\{\text{inference}\})}
	+ \lambda(\{\text{inference}\}) \cdot P_{\text{c}}(y)
\end{equation}

Where:
\begin{itemize}
	\item \( d = 0.75 \) is the discount factor,
	\item \( C(\{\left(\text{inference},1\right), \left(y,2\right)\}) \) is the count of the bigram,
	\item \( C(\{\text{inference}\}) \) is the count of the unigram "inference",
	\item \( \lambda(\{\text{inference}\}) = \frac{d}{C(\{\text{inference}\})} \cdot N_1^+(\{\text{inference}\}, \cdot) \) is the backoff weight,
	\item \( P_{\text{c}}(y) = \frac{N_1^+(\cdot, y)}{N} \) is the continuation probability,
	\item \( y \) is the candidate word that follows \textit{inference} in the text.
\end{itemize}

From the training text, we find that the word \textbf{"inference"} appears twice, and in both cases is followed by the word \textbf{"and"}. Applying the formula for the bigram \textbf{"inference and"}:

\[
\hat{p}_{Y \mid X}(Y = \text{and} \mid X = \{\text{inference}\}) = \frac{2 - 0.75}{2} + \frac{0.75}{2} \cdot \frac{2}{117}
= 0.625 + 0.375 \cdot 0.0171 \approx 0.631
\]

In fact, the word \textbf{"and"} is the only word that follows \textbf{"inference"} in the training data. Therefore, it is intuitive that \textbf{"and"} would be the predicted word even without performing the full calculation. However, in cases where multiple possible continuations exist, we select the word with the highest estimated probability.

It is also worth noting that we can introduce small random perturbations (e.g., Gaussian noise) to these probabilities to generate slightly different outputs for the same input context. This can add diversity and novelty to the predictions, making the generated text more dynamic and potentially more creative.
We now proceed to predict the next words sequentially using the smoothed bigram model. Starting with the initial phrase \textbf{inference and}, and applying the same Kneser-Ney smoothing method described earlier, the predicted third word is \textbf{deep}.
By continuing this process in an autoregressive manner to generate seven additional words, the resulting sequence becomes: \textbf{inference and deep learning including deep learning}. While this text lacks strong semantic coherence, it is important to note that the predictions are based on a very limited training corpus.
To evaluate the model’s performance on a larger dataset, the author applied the same approach to a corpus of approximately 50,000 words. In this setting, the generated sequences exhibited noticeably better coherence and semantic richness, highlighting the model’s potential when trained on more extensive linguistic data. 
It is important to note, however, that modern generative AI language models (LLMs) do not explicitly compute priors, likelihoods, or posteriors in the Bayesian sense. Instead, they learn a direct mapping from context to token probabilities using deep neural architectures. The model produces logits over the vocabulary and applies a softmax function to derive probability distributions for the next token.
Nevertheless, Bayes’ theorem provides a useful conceptual lens: the task of predicting missing words can be interpreted as reasoning under uncertainty, where the dependencies inherent in language structure implicitly guide the inference process. 
	
\section{Conclusion}
This paper has presented a unified probabilistic framework that connects estimation theory, statistical inference, and modern machine learning techniques—including deep learning and large language models. At its core, the challenge of learning from data is one of reasoning under uncertainty. Whether estimating hidden causes, learning input-output mappings, or modeling language structure, the same probabilistic principles recur. We showed that techniques like MLE, MAP, Bayesian classifiers, and attention-based architectures are not disparate methods, but rather specific manifestations of a shared mathematical foundation. This unified view demystifies complex AI models by rooting them in interpretable statistical logic. By examining both classical and modern paradigms, we illustrated how performance, interpretability, and generalization are governed by trade-offs tied to data complexity, model assumptions, and computational feasibility. This synthesis offers a principled guide for selecting or designing learning models across diverse domains, making the case that the future of machine intelligence lies in bridging rigorous mathematical foundations with scalable learning systems.
Finally, although the primary aim of this paper is to present the fundamental mathematical framework based on the uncertainty perspective for addressing problems typically tackled using artificial intelligence and machine learning, it is not intended to propose alternatives to the currently successful deep learning models. Nevertheless, due to the immense computational and energy demands of large-scale models—particularly large language models—revisiting foundational principles may inspire the development of novel architectures that are significantly less energy-intensive.


\begin{thebibliography}{1}
	
	\bibitem{ref1}
	E. Alpaydin, Introduction to Machine Learning, 4th Edition \emph{ MIT Press} 2020	
	\bibitem{ref2}
P. Moulin. and V. Veeravalli, 	Statistical Inference for Engineers and Data Scientists \emph{Cambridge Press}, 2018

\bibitem{ref3}
A. Vaswani. and N. Shazeer, N. Parmar, J. Uszkoreit, L. Jones, A. Gomes, L. kaiser, and I. Polosukhin, "Attention is All You Need",  \emph{Adanaces in Neural Information Processing Systems}, Vpl.30, 2017
\bibitem{ref4}
I. Goodfellow, J. Pouget-Abadie, M. Mirza, B. Xu, D. Warde-Farley, S. Ozair, A. Courville, and Y. Bengio, " Generative Adversarial Nets \emph{NeurlIPS}, 2014
\bibitem{ref5}
T. Mikolov, K. Chen, G. Corrado, and J. Dean, 	"Efficient Estimation of Word Representations in Vector Space" \emph{ICLR Workshop}, 2013
\bibitem{elmu}
M. Elmusrati, 	Modelling Stochastic Uncertainties: From Monte Carlo Simulations to Game Theory,  \emph{De Gruyter }, 2025
\bibitem{ref7}
LeCun, Y., Bengio, Y., and Hinton, G. "Deep learning",  Nature, 521(7553), 436–444, 2015
\bibitem{ref8}
Brown, T. et al.  "Language Models are Few-Shot Learners", NeurIPS. 2020
\bibitem{ref9}
Ghahramani, Z.  "Probabilistic machine learning and artificial intelligence". Nature, 521(7553), 452–459, 2005

\bibitem{ref10}
Bishop, C. M, Pattern Recognition and Machine Learning, Springer, 2006.

\bibitem{ref11}
S. Chen and J. Goodman, "An empirical study of smoothing techniques for language", Computer Speech and Language, Volume 13, Issue 4, October 1999
\end{thebibliography}
\end{document}